\documentclass[10pt,twocolumn,letterpaper]{article}

\usepackage{cvpr}
\usepackage{times}
\usepackage{epsfig}
\usepackage{graphicx}
\usepackage{amsmath}
\usepackage{amssymb}

\usepackage{amsmath}
\usepackage{pifont}

\usepackage{caption} 
\usepackage{subcaption}
\usepackage{bm}
\usepackage{multirow}
\usepackage{bbm}
\usepackage{comment}
\usepackage[title]{appendix}

\makeatletter
\def\thickhline{%
  \noalign{\ifnum0=`}\fi\hrule \@height \thickarrayrulewidth \futurelet
   \reserved@a\@xthickhline}
\def\@xthickhline{\ifx\reserved@a\thickhline
               \vskip\doublerulesep
               \vskip-\thickarrayrulewidth
             \fi
      \ifnum0=`{\fi}}
\makeatother

\newlength{\thickarrayrulewidth}
\usepackage[pagebackref=true,breaklinks=true,letterpaper=true,colorlinks,bookmarks=false]{hyperref}

\cvprfinalcopy 

\def\cvprPaperID{5235} 

\ifcvprfinal\fi
\begin{document}

\title{Equalization Loss for Long-Tailed Object Recognition}

\author{Jingru Tan$^1$ \quad Changbao Wang$^2$ \quad Buyu Li$^3$ \quad Quanquan Li$^2$ \\ \quad  Wanli Ouyang$^4$ \quad Changqing Yin$^1$  \quad Junjie Yan$^2$  \\
$^1$Tongji University \quad $^2$SenseTime Research \quad $^3$The Chinese University of Hong Kong  \\  \quad $^4$The University of Sydney, SenseTime Computer Vision Research Group, Australia \\ 
{\tt\small \{tjr120,yinchangqing\}@tongji.edu.cn, \{wangchangbao,liquanquan,yanjunjie\}@sensetime.com} \\ 
{\tt\small byli@ee.cuhk.edu.hk,  wanli.ouyang@sydney.edu.au}
}

\maketitle
\thispagestyle{empty}

\begin{abstract}
   Object recognition techniques using convolutional neural networks (CNN) have achieved great success. However, state-of-the-art object detection methods still perform poorly on large vocabulary and long-tailed datasets, \eg LVIS. In this work, we analyze this problem from a novel perspective: each positive sample of one category can be seen as a negative sample for other categories, making the tail categories receive more discouraging gradients.
   Based on it, we propose a simple but effective loss, named equalization loss, to tackle the problem of long-tailed rare categories by simply ignoring those gradients for rare categories. The equalization loss protects the learning of rare categories from being at a disadvantage during the network parameter updating. Thus the model is capable of learning better discriminative features for objects of rare classes.
   Without any bells and whistles, our method achieves AP gains of 4.1\% and 4.8\% for the rare and common categories on the challenging LVIS benchmark, compared to the Mask R-CNN baseline. With the utilization of the effective equalization loss, we finally won the 1st place in the LVIS Challenge 2019. Code has been made available at: \url{https://github.com/tztztztztz/eql.detectron2}
\end{abstract}

\section{Introduction}

Recently, the computer vision community has witnessed the great success of object recognition because of the emerge of deep learning and convolutional neural networks (CNNs). Object recognition, which is a fundamental task in computer vision, plays a central role in many related tasks, such as re-identification, human pose estimation and object tracking.

\begin{figure}[t]
   \centering 
   \includegraphics[width=\linewidth]{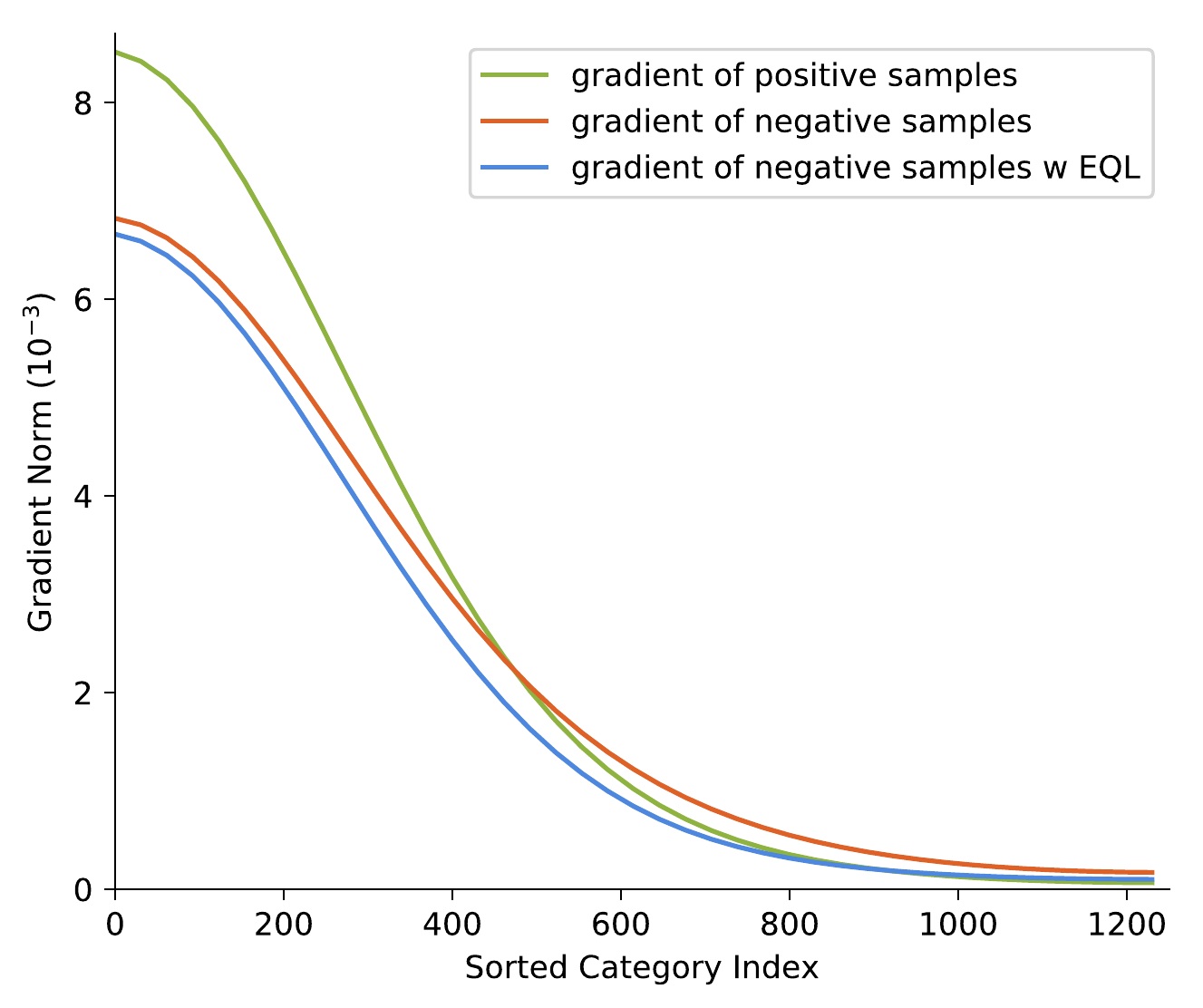}
   \caption{The overall gradient analysis on positive and negative samples. We collect the average $L_{2}$ norm of gradient of weights in the last classifier layer. Categories' indices are sorted by their instance counts. Note that for one category, proposals of all the other categories and the background are negative samples for it.}
   \label{fig:effect_of_eql_gradient}
\end{figure}

Today, most datasets for general object recognition, \eg Pascal VOC~\cite{everingham2010pascal} and COCO \cite{lin2014coco}, mainly collect frequently seen categories, with a large number of annotations for each class. However, when it comes to more practical scenarios, a large vocabulary dataset with a long-tailed distribution of category frequency (\eg LVIS~\cite{gupta2019lvis}) is inevitable. 
The problem of the long-tailed distribution of the categories is a great challenge to the learning of object detection models, especially for the rare categories (categories with very few samples). Note that for one category, all the samples of other categories including the background are regarded as negative samples. So the rare categories can be easily overwhelmed by the majority categories (categories with a large number of samples) during training and are inclined to be predicted as negatives. Thus the conventional object detectors trained on such an extremely unbalanced dataset suffer a great decline.

Most of the previous works consider the influence of the long-tailed category distribution problem as an imbalance of batch sampling during training, and they handle the problem mainly by designing specialized sampling strategies~\cite{chawla2002smote,han2005borderline,mahajan2018exploring,shen2016relay}. Other works introduce specialized loss formulations to cope with the problem of positive-negative sample imbalance \cite{lin2017focal,li2019gradient}. But they focus on the imbalance between foreground and background samples so that the severe imbalance among different foreground categories remains a challenging problem. 

\begin{figure}
   \centering 
   \includegraphics[width=\linewidth]{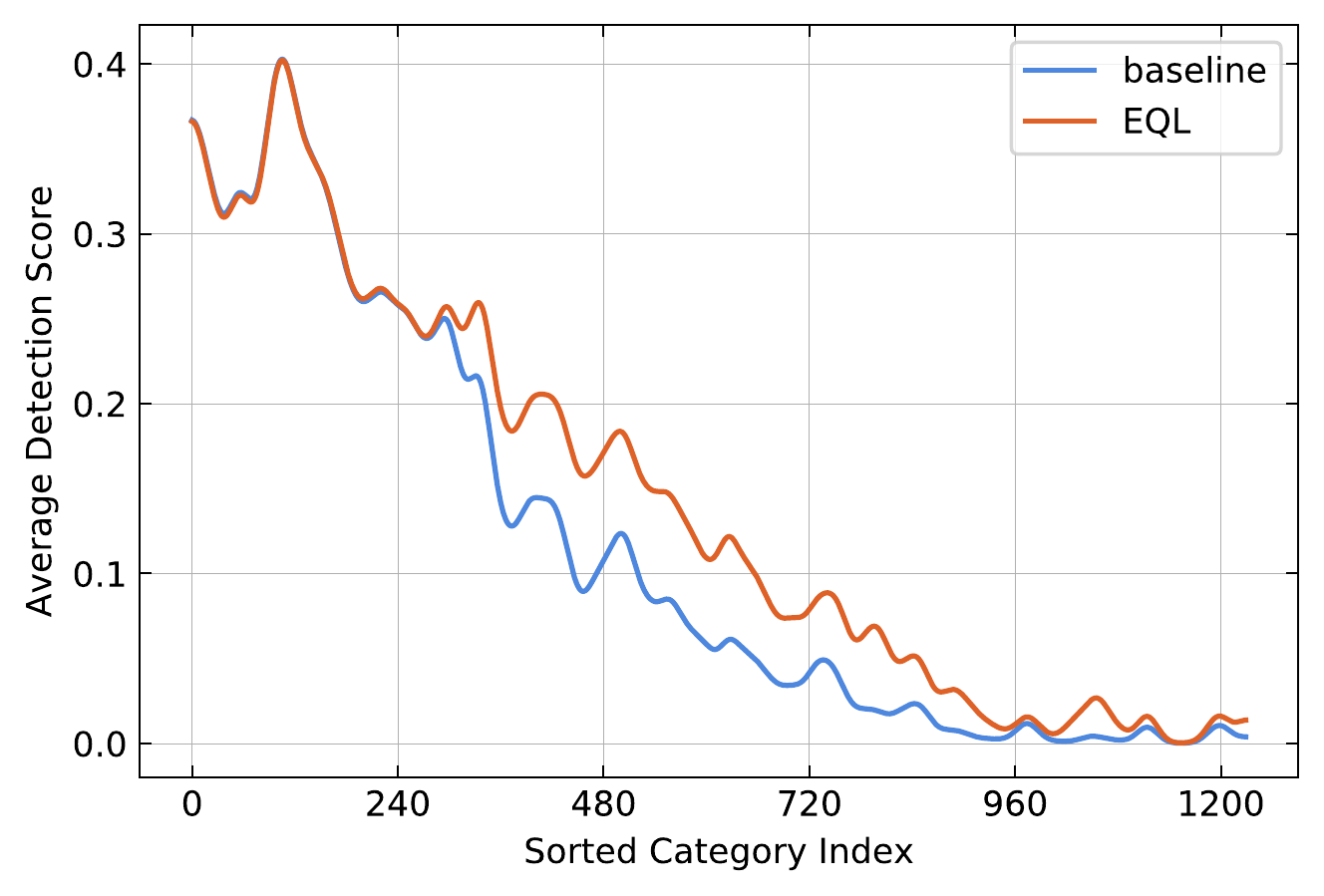}
   \caption{The predicted probabilities with EQL. The x-axis is the category index ordered by the instance number of each category, and the y-axis is the average predicted probability for positive proposals of each category. 
   }
   \label{fig:eql_effect_score}
\end{figure}

In this work, we focus on the problem of extremely imbalanced frequencies among different foreground categories and propose a novel perspective to analyze the effect of it. 
As illustrated in Figure \ref{fig:effect_of_eql_gradient}, the green and orange curves represent the average norms of gradients contributed by positive and negative samples respectively.  We can see that for the frequent categories, the positive gradient has a larger impact than the negative gradient on average, but for the rare categories, the status is just the opposite. 
To put it further, the commonly used loss functions in classification tasks, \eg softmax cross-entropy and sigmoid cross-entropy, have a suppression effect on the classes that are not the ground-truth one. When a sample of a certain class is utilized for training, the parameters of the prediction of the other classes will receive discouraging gradients which lead them to predict low probabilities. Since the objects of the rare categories hardly occur, the predictors for these classes are overwhelmed by the discouraging gradients during network parameters updating.

To address this problem, we propose a novel loss function, \textit{equalization loss (EQL)}. In general, we introduce a weight term for each class of each sample, which mainly reduces the influence of negative samples for the rare categories. The complete formulation of equalization loss is presented in Section \ref{sec:eql}. With the equalization loss, the average gradient norm of negative samples decrease as shown in Figure \ref{fig:effect_of_eql_gradient} (the blue curve). And a simple visualization of the effect of EQL is shown in Figure \ref{fig:eql_effect_score}, which illustrates the average predicted probabilities for the positive proposals of each category with (the red curve) and without (the blue curve) equalization loss. It can be seen that EQL significantly improves the performance on rare categories without harming the accuracy of frequent categories.
With the proposed EQL, categories of different frequencies are brought to a more equal status during network parameter updating, and the trained model is able to distinguish objects of the rare categories more accurately.

Extensive experiments on several unbalanced datasets, \eg Open Images \cite{kuznetsova2018open} and LVIS \cite{gupta2019lvis}, demonstrate the effectiveness of our method. We also verify our method on other tasks, like image classification.

Our key contributions can be summarized as follows: (1) We propose a novel perspective to analyze the long-tailed problem: the suppression on rare categories during learning caused by the inter-class competition, which explains the poor performance of rare categories on long-tailed datasets. Based on this perspective, a novel loss function, equalization loss is proposed, which alleviates the effect of the overwhelmed discouraging gradients during learning by introducing an ignoring strategy. (2) We present extensive experiments over different datasets and tasks, like object detection, instance segmentation and image classification. All experiments demonstrate the strength of our method, which brings a large performance boosting over common classification loss functions. Equipped with our equalization loss, we achieved the 1st place in the LVIS Challenge 2019.


\section{Related Works}

We first revisit common objection detection and instance segmentation. Then we introduce re-sampling, cost-sensitive re-weighting, and feature manipulation methods that are widely used to alleviate the class-unbalanced problem in long-tailed datasets.

\textbf{Object Detection and Instance Segmentation.} There are two mainstream frameworks for objection detection:
single-stage detector \cite{liu2016ssd, redmon2016you, lin2017focal} and two-stage detector \cite{girshick2014rich, girshick2015fast, ren2015faster, lin2017feature, lu2019grid}. 
While single-stage detectors achieve higher speed, most of state-of-the-art detectors follow the two-stage regime for better performance. The popular Mask R-CNN \cite{he2017mask}, which extends a mask head in the typical two-stage detector, provided promising results on many instance segmentation benchmarks. 
Mask Scoring R-CNN \cite{huang2019mask} introduced an extra mask score head to align the mask's score and quality. And Cascade Mask R-CNN \cite{cai2018cascade} and HTC \cite{chen2019hybrid} further improved the performance by predicting the mask in a cascade manner.

\textbf{Re-sampling Methods.} One of the commonly used methods in re-sampling is oversampling \cite{chawla2002smote, han2005borderline, mahajan2018exploring}, which randomly samples more training data from the minority classes, to tackle the unbalanced class distribution. Class-aware sampling \cite{shen2016relay}, also called class-balanced sampling, is a typical technique of oversampling, which first samples a category and then an image uniformly that contains the sampled category. 
While oversampling methods achieve significant improvement for under-represented classes, they come with a high potential risk of overfitting. On the opposite of oversampling, the main idea of under-sampling \cite{drummond2003c4} is to remove some available data from frequent classes to make the data distribution more balanced. However, the under-sampling is infeasible in extreme long-tailed datasets, since the imbalance ratio between the head class and tail class are extremely large. Recently, \cite{kang2019decoupling} proposed a decoupling training schema, which first learns the representations and classifier jointly, then obtains a balanced classifier by re-training the classifier with class-balanced sampling. Our method helps the model learn better representations for tail classes, so it could be complementary to the decoupling training schema.

\textbf{Re-weighting Methods.} The basic idea of re-weighting methods is to assign weights for different training samples. In an unbalanced dataset, an intuitive strategy is to weight samples based on the inverse of class frequency \cite{wang2017learning,huang2016learning} or use a smoothed version, inverse square root of class frequency \cite{mikolov2013distributed}. Besides methods mentioned above which adjust the weight on class level, there are other studies focus on re-weighting on sample level. \cite{lin2017focal, li2019gradient} make the neural network to be cost-sensitive by increasing the weight for hard samples and decreasing the weight for easy samples, which can be seen as online versions of hard example mining technique \cite{shrivastava2016training}. Recently, Meta-Weight-Net \cite{shu2019meta} learns an explicit mapping for sample re-weighting. Different from the works above, we focus on the imbalance problem among different foreground categories. We propose a new perspective that the large number of negative gradients from frequent categories severely suppress the learning of rare categories during training. And we propose a new loss function to tackle this problem, which is applied to the sample level and class level simultaneously.

\textbf{Feature Manipulation.} There are also some works operating on the feature representations directly. Range Loss \cite{zhang2017range} enlarges inter-classes distance and reduces intra-classes variations simultaneously. \cite{yin2019feature} augments the feature space of tail classes by transferring the feature variance of regular classes that have sufficient training samples. \cite{liu2019large} transfers the semantic feature representation from head to tail categories by adopting a memory module.
However, designing those modules or methods is not a trivial task and makes the model harder to train. In contrast, our method is simpler and does not access the representation directly.

\section{Equalization Loss} \label{sec:eql}
The central goal of our equalization loss is to alleviate the category quantity distribution imbalance problem for each category in a long-tailed class distribution. We start by revisiting conventional loss functions for classification, namely softmax cross-entropy and sigmoid cross-entropy.

\subsection{Review of Cross-Entropy Loss}

\textbf{Softmax Cross-Entropy} derives a multinomial distribution $\bm{p}$ over each category from the network outputs $\bm{z}$, and then computes the cross-entropy between the estimated distribution $\bm{p}$ and ground-truth distribution $\bm{y}$. The softmax cross-entropy loss $L_{SCE}$ can be formulated as:

\begin{equation}
   L_{SCE} = -\sum_{j=1}^{C} y_{j} \log(p_{j})
\end{equation}


\noindent and $C$ is the number of categories. Here, $\bm{p}$ is calculated by $Softmax(\bm{z})$. Note that the $C$ categories include an extra class for background. In practice, $\bm{y}$ uses one-hot representation, and we have $\sum_{j=1}^{C} y_{j}=1$. Formally, for the ground truth category $c$ of a sample,
\begin{equation}
   y_j = 
   \begin{cases}
      1 & \text{if } j = c \\
      0 & \text{otherwise}
  \end{cases} 
\end{equation}

\textbf{Sigmoid Cross-Entropy} estimates the probability of each category independently using $C$ sigmoid loss functions. The ground truth label $y_j$ only represents a binary distribution for category $j$. Usually, an extra category for background is not included. Instead, $y_j = 0$ will be set for all the categories when a proposal belongs to the background. So the sigmoid cross-entropy loss can be formulated as:
\begin{equation}
   L_{BCE} = -\sum_j^C \log(\hat{p_j})
\end{equation}

\noindent where
\begin{equation}
   \hat{p_j} =
   \begin{cases}
       p_j & \text{if } y_j = 1 \\
       1-p_j  & \text{otherwise}
   \end{cases} 
\end{equation}


\noindent Where $p_j$ is calculated by $\sigma(z_j)$. The derivative of the $L_{BCE}$ and $L_{SCE}$ with respect to network's output $\bm{z}$ in sigmoid cross entropy shares the same formulation: 

\begin{equation}
   \frac{\partial L_{cls}}{\partial z_{j}} = 
   \begin{cases}
      p_{j} - 1 & \text{if } y_j = 1 \\
      p_{j}              & \text{otherwise}
  \end{cases}
  \label{eq:grad}
\end{equation}


In softmax cross-entropy and sigmoid cross-entropy, we notice that for a foreground sample of category $c$, it can be regarded as a negative sample for any other category $j$. So the category $j$ will receive a discouraging gradient $p_{j}$ for model updating, which will lead the network to predict low probability for category $j$. If $j$ is a rare category, the discouraging gradients will occur much more frequently than encouraging gradients during the iterations of optimization. The accumulated gradients will have a non-negligible impact on that category. Finally, even positive samples for category $j$ might get a relatively low probability from the network.

\subsection{Equalization Loss Formulation}
When the quantity distribution of categories is fairly imbalanced, \eg in a long-tailed dataset, the discouraging gradients from frequent categories have a remarkable impact on categories with scarce annotations. With commonly used cross-entropy losses, the learning of rare categories are easily suppressed.
To solve this problem, we propose the equalization loss, which ignores the gradient from samples of frequent categories for the rare categories. This loss function aims to make the network training more fair for each class, and we refer it as equalization loss.

Formally, we introduce a weight term $w$ to the original sigmoid cross-entropy loss function, and the equalization loss can be formulated as:
\begin{equation}
   L_{EQL} = -\sum_{j=1}^{C} w_{j} log(\hat{p_{j}})
   \label{eq:eql}
\end{equation}

\noindent For a region proposal $r$, we set $w$ with the following regulations:
\begin{equation}
   w_{j} = 1 - E(r) T_{\lambda}(f_j) (1 - y_j)
   \label{eq:eql_w}
\end{equation}

\noindent In this equation, $E(r)$ outputs 1 when $r$ is a foreground region proposal and 0 when it belongs to background. And $f_{j}$ is the frequency of category $j$ in the dataset, which is computed by the image number of the class $j$ over the image number of the entire dataset. And $T_{\lambda}(x)$ is a threshold function which outputs 1 when $x < \lambda$ and 0 otherwise. $\lambda$ is utilized to distinguish tail categories from all other categories and Tail Ratio ($TR$) is used as the criterion to set the value of it. Formally, we define $TR$ by the following formula:

\begin{equation}
   TR(\lambda) = \frac{\sum_{j}^{C}T_\lambda(f_{j})N_{j}}{\sum_{j}^{C}N_{j}}
\end{equation}

\noindent where $N_{j}$ is the image number of category $j$. The settings of hyper-parameters of each part in Equation \ref{eq:eql_w} are studied in Section \ref{sec:ablation_study}.

In summary, there are two particular designs in equalization loss function: 1) We ignore the discouraging gradients of negative samples for rare categories whose quantity frequency is under a threshold. 2) We do not ignore the gradients of background samples. If all the negative samples for the rare categories are ignored, there will be no negative samples for them during training, and the learned model will predict a large number of false positives. 

\subsection{Extend to Image Classification}
Since softmax loss function is widely adopted in image classification, we also design a form of Softmax Equalization Loss following our main idea. Softmax equalization loss (SEQL) can be formulated as:

\begin{equation}
   L_{SEQL} = -\sum_{j=1}^{C} y_{j} \log(\tilde{p_{j}})
   \label{eq:seql}
\end{equation}

\noindent where
\begin{equation}
   \tilde{p_{j}} = \frac{e^{z_{j}}}{\sum_{k=1}^C\tilde{w_{k}}e^{z_{k}}}
   \label{eq:seql_p}
\end{equation}

\noindent and the weight term $w_{k}$ is computed by:
\begin{equation}
   \tilde{w_{k}} = 1 - \beta T_{\lambda}(f_k) (1 - y_k)
   \label{eq:seql_w}
\end{equation}
\noindent where $\beta$ is a random variable with a probability of $\gamma$ to be 1 and $1 - \gamma$ to be 0. 

Note that image classification is different from classification in object detection: each image belongs to a specific category, so there is no background category. Therefore, the weight term $\tilde{w_k}$ does not have the part $E(r)$ as in Equation \ref{eq:eql_w}. Therefore, we introduce $\beta$ to randomly maintain the gradient of negative samples. And the influence of $\gamma$ is studied in Section \ref{sec:cls}.

\begin{table*}
   \centering
   \begin{tabular}{l|c|c|c c c|c c c|c}
      & Backbone & EQL & AP & AP\textsubscript{50} & AP\textsubscript{75}& AP\textsubscript{\textit{r}} & AP\textsubscript{\textit{c}} & AP\textsubscript{\textit{f}}& AP\textsubscript{\textit{bbox}} \\
      \thickhline
       \multirow{2}*{Mask R-CNN} & \multirow{2}*{R-50-C4} & \ding{55} & 19.7 & 32.5 & 20.3 & 7.9 & 21.1 & 22.8 & 20.3 \\
       & & \ding{51} & 22.5 & 36.6 & 23.5 & 14.4 & 24.9 & 22.6 & 23.1 \\
       \thickhline
       \multirow{2}*{Mask R-CNN} & \multirow{2}*{R-101-C4} & \ding{55} & 21.8 & 35.6 & 22.7 & 10.5 & 23.4 & 24.2 & 22.9 \\
       & & \ding{51} & 24.1 & 38.7 & 25.6 & 15.8 & 26.8 & 24.1 & 25.6  \\
       \thickhline
       \multirow{2}*{Mask R-CNN} & \multirow{2}*{R-50-FPN} & \ding{55} & 20.1 & 32.7 & 21.2 & 7.2 & 19.9 & 25.4 & 20.5  \\
       & & \ding{51} & 22.8 & 36.0 & 24.4 & 11.3 & 24.7 & 25.1 & 23.3 \\
       \thickhline
       \multirow{2}*{Mask R-CNN} & \multirow{2}*{R-101-FPN} & \ding{55 }& 22.2 & 35.3 & 23.4 & 9.8 & 22.6 & 26.5 & 22.7  \\
       & & \ding{51} & 24.8 & 38.4 & 26.8 & 14.6 & 26.7 & 26.4 & 25.2  \\
       \thickhline
       \multirow{2}*{Cascade Mask R-CNN} & \multirow{2}*{R-50-FPN} & \ding{55} & 21.1 & 33.3 & 22.2 & 6.3 & 21.6 & 26.5 & 21.1  \\
       & & \ding{51} & 23.1 & 35.7 & 24.3 & 10.4 & 24.5 & 26.3 & 23.1  \\
       \thickhline
       \multirow{2}*{Cascade Mask R-CNN} & \multirow{2}*{R-101-FPN} & \ding{55} & 21.9 & 34.3 & 23.2 & 6.0 & 22.3 & 27.7 & 24.7  \\
       & & \ding{51} & 24.9 & 37.9 & 26.7 & 10.3 & 27.3 & 27.8 & 27.9  \\
       \thickhline
   \end{tabular}
   \caption{Results on different frameworks and models. All those models use class-agnostic mask prediction and are evaluated on LVIS v0.5 \texttt{val} set. AP is mask AP, and subscripts 'r', 'c' and 'f' stand for rare, common and frequent categories respectively. For equalization loss function, the $\lambda$ is set as $1.76 \times 10^{-3}$ to include all the rare and common categories.}
   \label{tab:results_on_frameworks}
\end{table*}

\section{Experiments on LVIS}

We conduct extensive experiments for equalization loss. In this section, we first present the implementation details and the main results on the LVIS dataset~\cite{gupta2019lvis} in Section~\ref{sec:imp} and Section~\ref{sec:eel}. Then we perform ablation studies to analyze different components of equalization loss in Section~\ref{sec:ablation_study}. In Section~\ref{sec:comparision}, we compare equalization loss with other methods. Details of LVIS Challenge 2019 will be introduced in Section~\ref{sec:lvischallenge}.

\subsection{LVIS Dataset}
\label{sec:lvis}

LVIS is a large vocabulary dataset for instance segmentation, which contains 1230 categories in current version v0.5. In LVIS, categories are divided into three groups according to the number of images that contains those categories: rare (1-10 images), common (11-100), and frequent ($>$100). We train our model on 57k \texttt{train} images and evaluate it on 5k \texttt{val} set. We also report our results on 20k \texttt{test} images. The evaluation metric is AP across IoU threshold from 0.5 to 0.95 over all categories. Different from COCO evaluation process, since LVIS is a sparse annotated dataset, detection results of categories that are not listed in the image level labels will not be evaluated.

\subsection{Implementation Details}
\label{sec:imp}

We implement standard Mask R-CNN \cite{he2017mask} equipped with FPN \cite{lin2017feature} as our baseline model. Training images are resized such that its shorter edge is 800 pixels while the longer edge is no more than 1333. No other augmentation is used except horizontal flipping. In the first stage, RPN samples 256 anchors with a 1:1 ratio between the foreground and background, and then 512 proposals are sampled per image with 1:3 foreground-background ratio for the second stage. We use 16 GPUs with a total batch size 32 for training. Our model is optimized by stochastic gradient descent (SGD) with momentum 0.9 and weight decay 0.0001 for 25 epochs, with an initial learning rate 0.04, which is decayed to 0.004 and 0.0004 at 16 epoch and 22 epoch respectively. Though class-specific mask prediction achieves better performance, we adopt a class-agnostic regime in our method due to the huge memory and computation cost for the large scale categories. Following~\cite{gupta2019lvis}, the threshold of prediction score is reduced from 0.05 to 0.0, and we keep the top 300 bounding boxes as prediction results. We make a small modification when EQL is applied on LVIS. Since for each image LVIS provide additional image-level annotations of which categories are in that image (positive category set) and which categories are not in it (negative category set), categories in EQL will not be ignored if they are in the positive category set or negative category set of that image, \ie the weight term of Equation~\ref{eq:eql_w} will be 1 for those categories, even if they are rare ones.

\subsection{Effectiveness of Equalization Loss}
\label{sec:eel}

Table \ref{tab:results_on_frameworks} demonstrates the effectiveness of equalization loss function over different backbones and frameworks. Besides Mask R-CNN, we also apply equalization loss on Cascade Mask R-CNN~\cite{cai2018cascade}. Our method achieves consistent improvement on all those models. As we can see from the table, the improvement mainly comes from the rare and common categories, indicating the effectiveness of our method on categories of the long-tailed distribution.

\begin{table}
   \centering
   \setlength\tabcolsep{5pt}
   \begin{tabular}{c c| c c c c|c}
         $\lambda (10^{-3})$ & TR(\%) & AP & AP\textsubscript{\textit{r}} & AP\textsubscript{\textit{c}} & AP\textsubscript{\textit{f}} & AP\textsubscript{\textit{bbox}} \\
         \thickhline
         0 & 0 & 20.1 & 7.2 & 19.9 & 25.4 & 20.5 \\
         \thickhline
         0.176($\lambda_{\textit{r}}$) & 0.93 & 20.8 & \textbf{11.7} & 20.2 & 25.2 & 20.8 \\
         0.5 & 3.14 &  22.0 & 11.2 & 22.8 & 25.2 & 22.4 \\
         0.8 & 4.88 & 22.3 & 11.4 & 23.4 & 25.3 & 23.0 \\
         1.5 & 7.82 & 22.8 & 11.0 & 24.5 & 25.5 & 23.0 \\
         1.76($\lambda_{\textit{c}}$) & 9.08 & \textbf{22.8} & 11.3 & \textbf{24.7} & 25.1 & \textbf{23.3} \\
         2.0 & 9.83 & 22.7 & 11.3 & 24.3 & 25.3 & 23.2 \\
         3.0 & 13.12 & 22.5 & 11.0 & 24.0 & 25.3 & 23.1 \\
         5.0 & 18.17 & 22.4 & 10.0 & 23.6 & \textbf{25.7} & 23.0 \\
         \thickhline
   \end{tabular}
   \caption{Ablation study for different $\lambda$. $\lambda_{\textit{r}}$ is about $1.76 \times 10^{-4}$, which exactly includes all rare categories. $\lambda_{\textit{c}}$ is about $ 1.76 \times 10^{-3}$, which exactly includes all rare and common categories. When $\lambda$ is 0, our equalization loss degenerates to sigmoid cross-entropy.}
   \label{tab:ablation_lambda}
\end{table}

\subsection{Ablation Studies}
\label{sec:ablation_study}
To better analyze equalization loss, we conduct several ablation studies. For all experiments we use ResNet-50 Mask R-CNN.

\noindent \textbf{Frequency Threshold $\lambda$:} The influence of different $\lambda$ is shown in Table~\ref{tab:ablation_lambda}. We perform experiments of changing $\lambda$ from $1.76 \times 10^{-4}$, which exactly split rare categories from all categories, to a broad range.  We empirically find the proper $\lambda$ locating in the space when $TR(\lambda)$ ranges from 2\% to 10\%. Results in Table~\ref{tab:ablation_lambda} shows that significant improvement of overall AP as $\lambda$ increases to include more tail categories. Meanwhile, the performance tends to degenerate when  $\lambda$ increases to include frequent categories. One advantage of equalization loss is that it has negligible effect on categories whose frequency is larger than a given $\lambda$. When $\lambda = \lambda_{r}$, $AP_r$ improves significantly with marginal influence to $AP_c$ and $AP_f$. And when $\lambda = \lambda_{c}$, $AP_r$ and $AP_c$ improve a lot while $AP_f$ only degenerates slightly. We set $\lambda$ to $\lambda_{c}$ in all our experiments. 

\begin{figure}
   \centering
   \includegraphics[width=\linewidth]{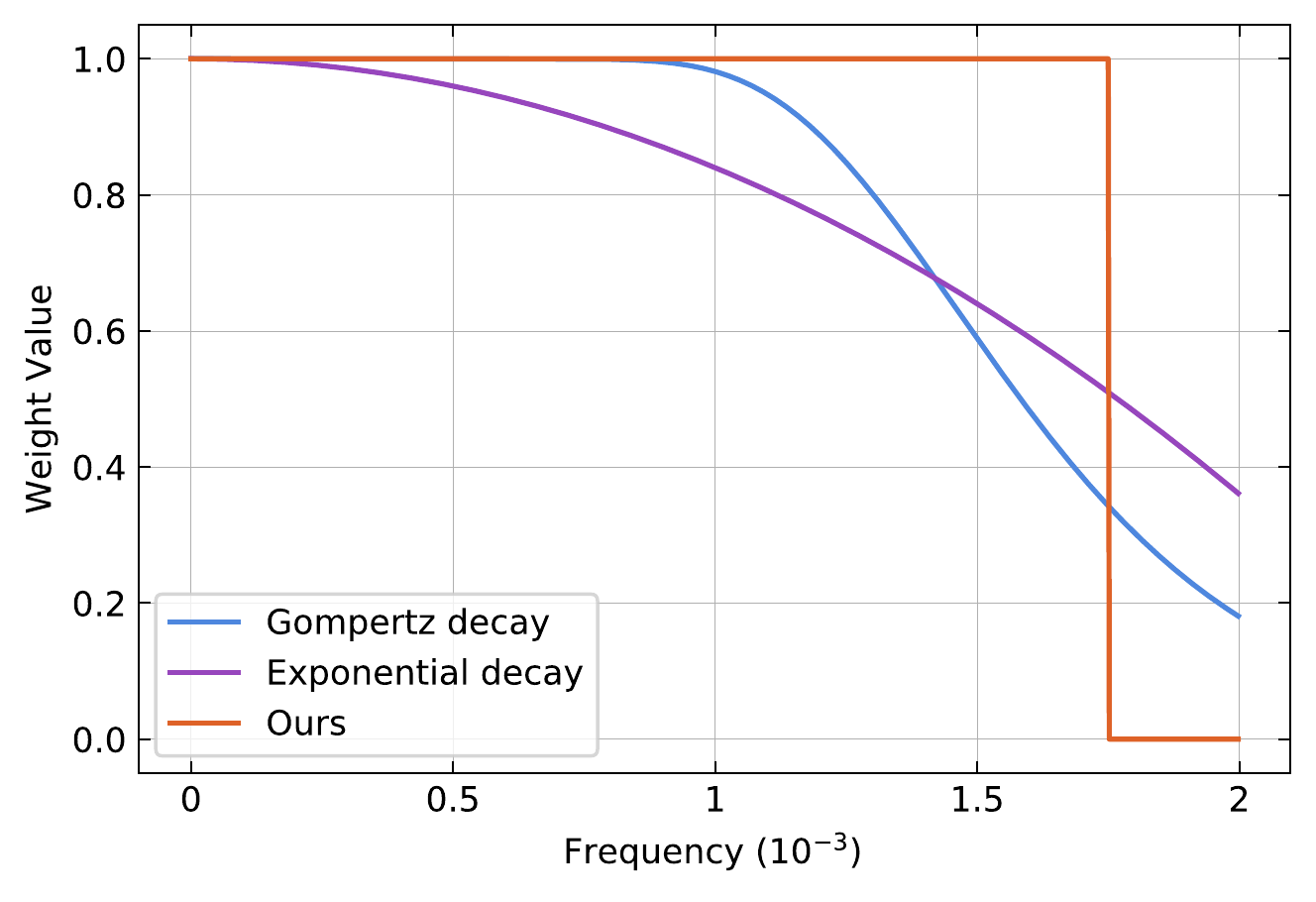}
   \caption{Illustration of different design for threshold function $T_\lambda(f)$.}
   \label{fig:threshold_function}
\end{figure}

\noindent \textbf{Threshold Function $T_\lambda(f)$:}
In Equation \ref{eq:eql_w}, we use $T_\lambda(f_j)$ to compute the weight of category $j$ for a given proposal. Except for the proposed threshold function, $T_\lambda(f)$ can have other forms to calculate the weight for the categories with frequency under the threshold. As illustrated in Figure \ref{fig:threshold_function}, we present and compare with another two designs: (1) Exponential decay function $ y = 1 - (af)^{n}$, which computes the weight according to the power of category frequency. (2) Gompertz decay function $ y = 1 - a e^{-be^{-cf}}$, which decays smoothly at the beginning and then decreases more steeply. We run multiple experiments for Exponential decay function and Gompertz decay function with different hyper-parameters and report the best results. The best hyper-parameter settings for Exponential decay function is $a = 400 $ and $n = 2$ and for Gompertz decay function $a = 1, b = 80, c = 3000 $. Table \ref{tab:ablation_weight_value} shows that all of the three designs achieve fairly similar results, while both exponential decay and Gompertz decay function introduce more hyper-parameters to fit the design. Therefore, we use the threshold function in our method for its simpler format with less hyper-parameters and better performance.

\begin{table}
   \centering
   \setlength\tabcolsep{4.5pt}
   \begin{tabular}{l|c c c c|c}
        & AP & AP\textsubscript{\textit {r}} & AP\textsubscript{\textit{c}} &  AP\textsubscript{\textit{f}} & AP\textsubscript {\textit{bbox}} \\
        \thickhline
       Exponential decay & 22.3 & 10.4 & 24.0 & 25.0 & 22.8 \\
       \thickhline
       Gompertz decay & 22.7 & 11.0 & 24.5 & 25.1 & 23.2 \\
       \thickhline
        Ours & \textbf{22.8} & \textbf{11.3} & \textbf{24.7} & \textbf{25.1} & \textbf{23.3} \\
       \thickhline
   \end{tabular}
   \caption{Ablation study for threshold function $T_\lambda(f)$. For a fair comparison, we compare the performance with their best hyper-parameters in multiple experiments.}
   \label{tab:ablation_weight_value}
\end{table}

\begin{table}
   \centering
   \setlength\tabcolsep{4pt}
   \begin{tabular}{c|c c c c|c }
         $E(r)$ &  AP & AP\textsubscript{\textit {r}} & AP\textsubscript{\textit{c}} & AP\textsubscript{\textit{f}} & AP\textsubscript{\textit{bbox}} \\
        \thickhline
        \ding{55} & 22.2 & \textbf{12.5} & 24.7 & 23.1 & 22.7 \\
        \ding{51} & \textbf{22.8} & 11.3 & 24.7 & \textbf{25.1} & \textbf{23.3} \\
       \thickhline
   \end{tabular}
   \caption{Ablation study of Excluding Function $E(r)$. The top row is the results without using the term $E(r)$, and the bottom row is the results with it.}
   \label{tab:ablation_ignore_negative}
\end{table}

\noindent \textbf{Excluding Function $E(r)$:} Table~\ref{tab:ablation_ignore_negative} shows the experiment results for EQL with or without $E(r)$. EQL without $E(r)$ means removing $E(r)$ from Equation \ref{eq:eql_w}, which will treat the foreground and background the same way. EQL with $E(r)$ means equalization loss only affects foreground proposals, as defined in Equation \ref{eq:eql_w}. Experiment results demonstrate the importance of $E(r)$. As we can see from the table, with $E(r)$, EQL achieves 0.6 points AP gain compared with EQL without $E(r)$. If $E(r)$ is discarded, although $AP_r$ has an increase, $AP_f$ drops dramatically, which causes the overall AP decline. 

\begin{figure}
   \centering 
   \includegraphics[width=0.9\linewidth]{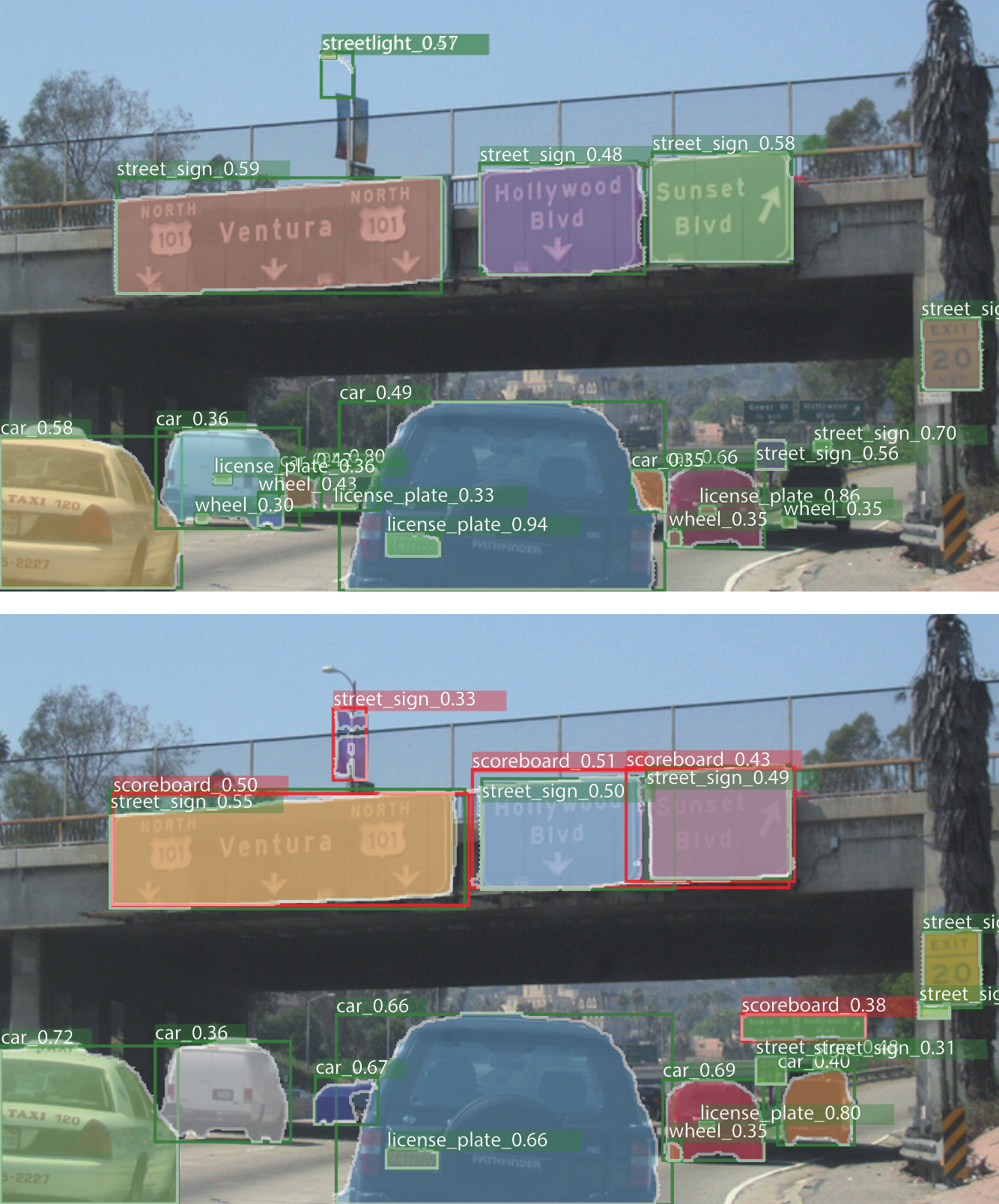}
   \caption{Illustration of the effect of excluding function $E(r)$. The upper and lower images correspond to using and removing $E(r)$ respectively. 
   The true positives are drawn in green while false positives in red. We visualize the results with scores higher than 0.3 for better visualization.}
   \label{fig:with_w_g}
\end{figure}

It is worth to notice that if we don't use $E(r)$, a large number of background proposals will be also ignored for rare and common categories, and the insufficient supervision from background proposals will cause extensive false positives. We visualize the detection results of an example image, which is shown in Figure~\ref{fig:with_w_g}. Without $E(r)$, more false positives are introduced, which are shown in red color. Both analysis and illustration above indicate that $AP_r$ should decrease without $E(r)$, which is contradictory with the experiment results in Table~\ref{tab:ablation_ignore_negative}. The reason is that according to LVIS evaluation protocol, if it is not sure whether category $j$ is in or not in image $I$, all the false positives of category $j$ will be ignored in image $I$. If category $j$ is rare, the increased false positives are mostly ignored, which alleviates their influence. But the simultaneously increased true positives bring a direct increase in $AP_r$.

\begin{table*}
   \centering
   \begin{tabular}{l|c c c|c c c|c c c|c}
      & AP & AP\textsubscript{50} & AP\textsubscript{75}& AP\textsubscript{\textit{r}} & AP\textsubscript{\textit{c}} & AP\textsubscript{\textit{f}} & AP\textsubscript{\textit{S}} & AP\textsubscript{\textit{M}} & AP\textsubscript{\textit{L}} & AP\textsubscript{\textit{bbox}} \\
      \thickhline
       Sigmoid Loss & 20.1 & 32.7 & 21.2 & 7.2 & 19.9 & 25.4 & 15.0 & 27.1 & 34.6 & 20.5 \\ 
       Softmax Loss & 20.2 & 32.6 & 21.3 & 4.5 & 20.8 & 25.6 & 15.5 & 27.7 & 34.4 & 20.7 \\
       \thickhline
       Class-aware Sampling \cite{shen2016relay}   & 18.5 & 31.1 & 18.9 & 7.3 & 19.3 & 21.9 & 13.3 & 24.3 & 30.5 & 18.4\\
       Repeat Factor Sampling \cite{gupta2019lvis} & 21.3 & 34.9 & 22.0 & \textbf{12.2} & 21.5 & 24.7 & 15.2 & 27.1 & 36.1 & 21.6 \\
       Class-balanced Loss \cite{cui2019class} & 20.9 & 33.8 & 22.2 & 8.2 & 21.2 & 25.7 & 15.6 & 28.1 & 35.3 & 21.0\\
       Focal Loss \cite{lin2017focal} & 21.0 & 34.2 & 22.1 & 9.3 & 21.0 & \textbf{25.8} & 15.6 & 27.8 & 35.4 & 21.9 \\
       \thickhline
       EQL(Ours) & \textbf{22.8} & \textbf{36.0} & \textbf{24.4} & 11.3 & \textbf{24.7} & 25.1 & \textbf{16.3} & \textbf{29.7} & \textbf{38.2} & \textbf{23.3} \\
       \thickhline
   \end{tabular}
      \caption{Comparison with other methods on LVIS v0.5 \texttt{val} set. All experiments are performed based on ResNet-50 Mask R-CNN.
      }
   \label{tab:compare_other_loss}
\end{table*}


\subsection{Comparison with Other Methods}
\label{sec:comparision}
Table \ref{tab:compare_other_loss} presents the comparison with other methods that are widely adopted to tackle the class imbalance problem. According to the table, re-sampling methods improve $AP_r$ and $AP_c$ at the sacrifice of $AP_f$, while re-weighting methods bring consistent gains on all categories but the overall improvement is trivial.
The equalization loss improves $AP_r$ and $AP_c$ significantly with slight effect on $AP_f$, surpassing all other approaches.

\subsection{LVIS Challenge 2019}
\label{sec:lvischallenge}

With the help of the equalization loss, we finally won the 1st place on LVIS challenge held on COCO and Mapillary Joint Recognition Challenge 2019. Combined with other enhancements, like larger backbone \cite{hu2018squeeze,xie2017aggregated}, deformable convolution~\cite{dai2017deformable}, synchronized batch normalization~\cite{peng2018megdet}, and extra data, our method achieves a 28.9 mask AP on LVIS v0.5 \texttt{test} set, outperforming the ResNeXt-101 Mask R-CNN baseline (20.1\%) by 8.4\%.  More details about our solution of challenge are described in Appendix A.

\section{Experiments on Open Images Detection}
\label{sec:openimage}

Open Image dataset v5 is a large dataset of ~9M images annotated with image-level labels and bounding boxes. In our experiments, we use the split of data and the subset of the categories of the competition 2019 for object detection track (OID19). The \texttt{train} set of OID19 contains 12.2M bounding boxes over 500 categories on 1.7M images, and the \texttt{val} contains about 10k images. 
 
According to Table \ref{tab:openimage}, our method achieves a great improvement compared with standard sigmoid cross-entropy, outperforming class-aware sampling method by a significant margin.
To better understand the improvement of our methods, we group all the categories by their image number and report the performance of each group. We can see that our method has larger improvements on categories with fewer samples. Significant AP gains on the group of fewest 100 categories are achieved compared with sigmoid cross-entropy and class-aware sampling (2.6 and 10.88 points respectively).

\begin{table}
   \centering
   \small
   \setlength\tabcolsep{4pt}
   \begin{tabular}{l| c | c  c  c  c  c}
      Method & AP & AP\textsubscript{1}& AP\textsubscript{2} & AP\textsubscript{3}& AP\textsubscript{4} & AP\textsubscript{5}  \\
      \thickhline
      SGM & 48.13 & 59.86 & 51.24 & 49.31 & 46.51 & 33.72\\
      CAS \cite{shen2016relay}  & 56.50 & 64.44 & 59.30 & 59.74 & 57.02 & 42.00\\
      \thickhline
      EQL(Ours) & \textbf{57.83 }& \textbf{64.95} & \textbf{60.18} & \textbf{61.17} & \textbf{58.23} & \textbf{44.6 }\\
      \thickhline
   \end{tabular}
   
   \caption{Results on OID19 \texttt{val} set based on ResNet-50. \textbf{SGM} and \textbf{CAS} stand for sigmoid cross-entropy and class-aware sampling. We sort all the categories by their image number and divide them into 5 groups. $TR$ and $\lambda$ is 3\% and $3 \times 10^{-4}$ respectively.}
   \label{tab:openimage}
\end{table}

\section{Experiments on Image Classification}
\label{sec:cls}

To demonstrate the generalization ability of the equalization loss when transferring to other tasks. We also evaluate our method on two long-tailed image classification datasets, CIFAR-100-LT and ImageNet-LT.

\noindent \textbf{Datasets.} We follow exactly the same setting with ~\cite{cui2019class} to generate the CIFAR-100-LT with imbalance factor of 200 \footnote{\url{https://github.com/richardaecn/class-balanced-loss}}.  CIFAR-100-LT contains 9502 images in \texttt{train} set, with 500 images for the most frequent category and 2 images for the rarest category. CIFAR-100-LT shares the same test set of 10k images with original CIFAR-100. We report the top1 and top5 accuray.  ImageNet-LT \cite{liu2019large} is generated from ImageNet-2012 \cite{deng2009imagenet}, which contains 1000 categories with images number ranging from 1280 to 5 images for each category \footnote{\url{https://github.com/zhmiao/OpenLongTailRecognition-OLTR}}. There are 116k images for training and 50k images for testing. Different from CIFAR-100-LT, we additionally present accuracies of many shot, medium shot and few shot to measure the improvement on tail classes.

\begin{table}
   \centering
   \setlength\tabcolsep{4pt}
   \begin{tabular}{c | c  c | c c }
      \multicolumn{1}{c}{} & \multicolumn{2}{c}{$\lambda = 3.0 \times 10^{-3}$} &  \multicolumn{2}{c}{$\lambda = 5.0 \times 10^{-3}$} \\ 
     \thickhline
      $\gamma$ & Acc@top1 & Acc@top5 & Acc@top1 & Acc@top5\\
     \thickhline
     0   & 41.33 & 67.75 & 41.33 & 67.75 \\
     \thickhline
     0.75 & 42.08 & 70.03 & 42.26 & 69.95 \\
     0.9 &   43.12 & 71.50 & \textbf{43.74} & \textbf{71.42} \\
     0.95 &  \textbf{43.38} & \textbf{71.94} & 43.30 & 72.31 \\
     0.99 &   42.44 & 71.44 & 42.49 & 72.07 \\
     \thickhline
    \end{tabular}
    \caption{Varying $\gamma$ and $\lambda$ for SEQL. The accuracy is reported on CIFAR-100-LT \texttt{test} set. $\gamma=0$ means we use softmax loss function.}
    \label{tab:cifar_gamma}
\end{table}

\begin{table}
   \centering
   \begin{tabular}{l|c|c}
      Method & Acc@top1 & Acc@Top5 \\
      \thickhline 
      Focal Loss\textsuperscript{\textdagger}~\cite{lin2017focal} & 35.62 & - \\
      Class Balanced\textsuperscript{\textdagger}~\cite{cui2019class} & 36.23 & - \\
      Meta-Weight Net\textsuperscript{\textdagger}~\cite{{shu2019meta}} & 37.91 & - \\
      \thickhline
      SEQL(Ours) & \textbf{43.38} & \textbf{71.94}\\
      \thickhline
     \end{tabular}
     \caption{Results on CIFAR-100-LT \texttt{test} set based on ResNet-32 ~\cite{he2016deep}. We use $\gamma$ of 0.95 and $\lambda$ of $3.0 \times 10^{-3}$. \textdagger~means that the results are copied from origin paper \cite{cui2019class,shu2019meta}. Imbalanced factor is 200.}
   \label{tab:cifar100}
\end{table}


\noindent \textbf{Implementation Details.} For CIFAR-100-LT, we use Nesterov SGD with momentum 0.9 and weight decay 0.0001 for training. We use a total mini-batch size of 256 with 128 images per GPU. The model ResNet-32 is trained for 12.8K iterations with learning rate 0.2, which is then decayed by a factor of 0.1 at 6.4K and 9.6K iteration. Learning rate is increased gradually from 0.1 to 0.2 during the first 400 iterations. For data augmentation, we first follow the same setting as \cite{lee2015deeply,he2016deep}, then use autoAugment \cite{cubuk2019autoaugment} and Cutout \cite{devries2017improved}. In testing, we simply use the origin $32 \times 32$ images. For ImageNet-LT, we use a total mini-batch size of 1024 with 16 GPUs. We use ResNet-10 as our backbone like \cite{liu2019large}.The model is trained for 12K iterations with learning rate 0.4, which is divided by 10 at 3.4K, 6.8K, 10.2K iterations. A gradually warmup strategy \cite{goyal2017accurate} is also adopted to increase the learning rate from 0.1 to 0.4 during the first 500 iterations. We use random-resize-crop, color jitter and horizontal flipping as data augmentation. Training input size is $224 \times 224$. In testing, we resized the images to $256 \times 256$ then cropped a single view of $224 \times 224$ at the center.


 \noindent \textbf{Results on CIFAR-100-LT and ImageNet-LT.} We build a much stronger baseline on CIFAR-100-LT due to those augmentation techniques. As shown in Table \ref{tab:cifar_gamma}, our EQL still improves the strong baseline by a large margin of 2\%. And those improvements are come from classes with fewer training samples. As for ImageNet-LT, we also present ablation studies in Table \ref{tab:imaget_lt_gamma}. A wide range of values of $\gamma$ give consistent improvements over the softmanx loss baseline. As shown in Table \ref{tab:cifar100} and Table \ref{tab:imagenetlt}, our equalization loss surpasses prior state-of-the-art approaches significantly, which demonstrates that our method can be generalized to different tasks and datasets effectively.

\begin{table}
   \centering
   \setlength\tabcolsep{4pt}
   \begin{tabular}{c|c c c|c c}
      $\gamma$ & Many & Medium & Few & Acc@top1 & Acc@top5 \\
     \thickhline
     0    & \textbf{53.2} & 27.5 & 8.0 & 34.7 & 58.7 \\
     \thickhline
     0.5  & 52.5 & 28.7 & 9.8 & 35.2 & 59.7 \\
     0.75 & 52.1 & 30.7 & 11.6 & 36.2 & 60.8 \\
     0.9  & 49.4 & 32.3 & 14.5 & \textbf{36.4} & \textbf{61.1} \\
     0.95 & 46.5 & \textbf{32.8} & \textbf{16.4} & 35.8 & 60.7 \\
     \thickhline
    \end{tabular}
    \caption{Varying $\gamma$ for SEQL with $\lambda$ of $4.3 \times 10^{-4}$. The accuracy is reported on ImageNet-LT \texttt{test} set. When $\gamma$ is 0, the SEQL degenerates to softmax loss function. }
    \label{tab:imaget_lt_gamma}
\end{table}

\begin{table}
   \centering
   \begin{tabular}{l|c|c}
      Method & Acc@Top1 & Acc@Top5 \\
      \thickhline
      FSLwF\textsuperscript{\textdagger}~\cite{gidaris2018dynamic} & 28.4 & - \\
      Focal Loss\textsuperscript{\textdagger}~\cite{lin2017focal} & 30.5 & -  \\
      Lifted Loss\textsuperscript{\textdagger}~\cite{oh2016deep} & 30.8 & -  \\
      Range Loss\textsuperscript{\textdagger}~\cite{zhang2017range} & 30.7 & - \\
      OLTR\textsuperscript{\textdagger}~\cite{liu2019large} & 35.6 & - \\
      \thickhline
      SEQL(Ours) & \textbf{36.44} & \textbf{61.19} \\
      \thickhline
     \end{tabular}
   \caption{Results on ImageNet-LT \texttt{test} set based on ResNet-10 ~\cite{he2016deep}.
   The optimal $\gamma$ and $\lambda$ are 0.9 and $4.3 \times 10^{-4}$ respectively. \textdagger~means that the results are copied from origin paper \cite{liu2019large}}
   \label{tab:imagenetlt}
\end{table}

\section{Conclusion}


In this work, we analyze the severe inter-class competition problem in long-tailed datasets. We propose a novel equalization loss function to alleviate the effect of the overwhelmed discouraging gradients on tail categories. Our method is simple but effective, bringing a significant improvement over different frameworks and network architectures on challenging long-tailed object detection and image classification datasets. 

\textbf{Acknowledgment} Wanli Ouyang is supported by the Australian Research Council Grant  DP200103223.

\clearpage
\begin{appendices}

\begin{table}
    \centering
    \small
    \begin{tabular}{l|c c c c}
         Model & AP & AP\textsubscript{\textit {r}} & AP\textsubscript{\textit{c}} &  AP\textsubscript{\textit{f}}  \\
         \thickhline
         Challenge Baseline & 30.1 & 19.3 & 31.8 & 32.3 \\
         \thickhline
         +SE154 \cite{hu2018squeeze} & 30.8 & 19.7 & 32.2 & 33.4 \\
         +OpenImage Data & 31.4 & 21.5 & 33.1 & 33.3  \\
         +Multi-scale box testing & 32.3 & 20.5 & 34.7 & 34.2 \\
         +RS Ensemble+Expert Model & 35.1 & 24.8 & 37.5 & 36.3 \\
         +Multi-scale mask testing & \textbf{36.4} & \textbf{25.5}& \textbf{38.6} & \textbf{38.1} \\
         \thickhline
    \end{tabular}
    \caption{Experiment results of different tricks on LVIS v0.5 \texttt{val} set. RS Ensemble stands for Re-scoring Ensemble.}
    \label{tab:enhancement}
  \end{table}

\section{Details of LVIS Challenge 2019}
\label{sec:lvis_challenge_details}

With equalization loss, we ranked 1st entry on LVIS Challenge 2019. In this section, we will introduce details of the solution we used in the challenge. 

\noindent \textbf{External Data Exploiting} Since LVIS is not exhaustively annotated with all categories and the annotations for long-tailed categories are quite scarce, we utilize additional public datasets to enrich our training set.
First, we train a Mask R-CNN on COCO train2017 with 115k images and then fine-tune our model with equalization loss on LVIS. During fine-tuning, we leverage COCO annotations of bounding boxes as ignored regions to exclude background proposals during sampling. Moreover, we borrow $\sim$20k images from Open Images V5 which contains shared 110 categories with LVIS and use the bounding boxes annotations to train the model.

\noindent \textbf{Model Enhancements}
We achieve our challenge baseline by training ResNeXt-101-64x4d~\cite{xie2017aggregated} enhanced by deformable convolution~\cite{dai2017deformable} and synchronized batch normalization~\cite{peng2018megdet}, along with equalization loss, repeat factor sampling , multi-scale training and COCO exploiting, which lead to 30.1\% AP on LVIS v0.5 \texttt{val} set.
We apply multi-scale testing on both bounding box and segmentation results and the testing scale ranges from 600 to 1400 with step size of 200.
We train two expert models on train set of COCO 2017 and Open Images V5 respectively and then evaluate them on LVIS \texttt{val} set to collect the detection results of shared categories.
Though our method improves the performance of long-tailed categories a lot, the prediction scores for these categories tend to be smaller than frequent ones due to the lack of positive training samples, which leads to degeneration of $AP_r$ in ensemble.
To keep more results for rare and common categories, we employ a re-score ensemble approach via improving the scores of these categories.

Our road map is shown in \ref{tab:enhancement}.  
With those enhancements, we achieve 36.4 and 28.9 Mask AP on \texttt{val} and \texttt{test} set respectively which is demonstrated in Table \ref{tab:testset}.

\begin{table}[t]
    \centering
    \setlength\tabcolsep{5pt}
    \small
    \begin{tabular}{l|c c c | c c c}
      & AP & AP\textsubscript{50} & AP\textsubscript{75}& AP\textsubscript{\textit{r}} & AP\textsubscript{\textit{c}} & AP\textsubscript{\textit{f}} \\ 
         \thickhline
         Ours& \textbf{28.85} & \textbf{42.69} & \textbf{31.10} & \textbf{17.71} & \textbf{30.84} & 36.70\\
         \thickhline
         2nd place & 26.67 & 38.68 & 28.78 & 10.59 & 28.70 & \textbf{39.21} \\
         3rd place & 24.04 & 36.51 & 25.68 & 15.32 & 24.95 & 31.12 \\
         4th place & 22.75 & 34.29 & 24.17 & 11.67 & 23.51 & 32.33 \\
         Baseline~\cite{gupta2019lvis} & 20.49 & 32.33 & 21.57 & 9.80 & 21.05 & 30.00 \\
         \thickhline
    \end{tabular}
    \caption{Results reported on LVIS v0.5 20k \texttt{test} set. The equalization loss plays a important role to allow us to achieve the highest AP both on rare and common categories. This results can be accessed by \url{https://evalai.cloudcv.org/web/challenges/challenge-page/442/leaderboard/1226}}
    \label{tab:testset}
\end{table}

\end{appendices}

{\small
\bibliographystyle{ieee_fullname}
\bibliography{egbib}

\begin{thebibliography}{10}\itemsep=-1pt

\bibitem{cai2018cascade}
Zhaowei Cai and Nuno Vasconcelos.
\newblock Cascade r-cnn: Delving into high quality object detection.
\newblock In {\em Proceedings of the IEEE conference on computer vision and
  pattern recognition}, pages 6154--6162, 2018.

\bibitem{chawla2002smote}
Nitesh~V Chawla, Kevin~W Bowyer, Lawrence~O Hall, and W~Philip Kegelmeyer.
\newblock Smote: synthetic minority over-sampling technique.
\newblock {\em Journal of artificial intelligence research}, 16:321--357, 2002.

\bibitem{chen2019hybrid}
Kai Chen, Jiangmiao Pang, Jiaqi Wang, Yu Xiong, Xiaoxiao Li, Shuyang Sun,
  Wansen Feng, Ziwei Liu, Jianping Shi, Wanli Ouyang, et~al.
\newblock Hybrid task cascade for instance segmentation.
\newblock In {\em Proceedings of the IEEE Conference on Computer Vision and
  Pattern Recognition}, pages 4974--4983, 2019.

\bibitem{cubuk2019autoaugment}
Ekin~D Cubuk, Barret Zoph, Dandelion Mane, Vijay Vasudevan, and Quoc~V Le.
\newblock Autoaugment: Learning augmentation strategies from data.
\newblock In {\em Proceedings of the IEEE conference on computer vision and
  pattern recognition}, pages 113--123, 2019.

\bibitem{cui2019class}
Yin Cui, Menglin Jia, Tsung-Yi Lin, Yang Song, and Serge Belongie.
\newblock Class-balanced loss based on effective number of samples.
\newblock In {\em Proceedings of the IEEE Conference on Computer Vision and
  Pattern Recognition}, pages 9268--9277, 2019.

\bibitem{dai2017deformable}
Jifeng Dai, Haozhi Qi, Yuwen Xiong, Yi Li, Guodong Zhang, Han Hu, and Yichen
  Wei.
\newblock Deformable convolutional networks.
\newblock In {\em Proceedings of the IEEE international conference on computer
  vision}, pages 764--773, 2017.

\bibitem{deng2009imagenet}
Jia Deng, Wei Dong, Richard Socher, Li-Jia Li, Kai Li, and Li Fei-Fei.
\newblock Imagenet: A large-scale hierarchical image database.
\newblock In {\em 2009 IEEE conference on computer vision and pattern
  recognition}, pages 248--255. Ieee, 2009.

\bibitem{devries2017improved}
Terrance DeVries and Graham~W Taylor.
\newblock Improved regularization of convolutional neural networks with cutout.
\newblock {\em arXiv preprint arXiv:1708.04552}, 2017.

\bibitem{drummond2003c4}
Chris Drummond, Robert~C Holte, et~al.
\newblock C4. 5, class imbalance, and cost sensitivity: why under-sampling
  beats over-sampling.
\newblock In {\em Workshop on learning from imbalanced datasets II}, volume~11,
  pages 1--8. Citeseer, 2003.

\bibitem{everingham2010pascal}
Mark Everingham, Luc Van~Gool, Christopher~KI Williams, John Winn, and Andrew
  Zisserman.
\newblock The pascal visual object classes (voc) challenge.
\newblock {\em International journal of computer vision}, 88(2):303--338, 2010.

\bibitem{gidaris2018dynamic}
Spyros Gidaris and Nikos Komodakis.
\newblock Dynamic few-shot visual learning without forgetting.
\newblock In {\em Proceedings of the IEEE Conference on Computer Vision and
  Pattern Recognition}, pages 4367--4375, 2018.

\bibitem{girshick2015fast}
Ross Girshick.
\newblock Fast r-cnn.
\newblock In {\em Proceedings of the IEEE international conference on computer
  vision}, pages 1440--1448, 2015.

\bibitem{girshick2014rich}
Ross Girshick, Jeff Donahue, Trevor Darrell, and Jitendra Malik.
\newblock Rich feature hierarchies for accurate object detection and semantic
  segmentation.
\newblock In {\em Proceedings of the IEEE conference on computer vision and
  pattern recognition}, pages 580--587, 2014.

\bibitem{goyal2017accurate}
Priya Goyal, Piotr Doll{\'a}r, Ross Girshick, Pieter Noordhuis, Lukasz
  Wesolowski, Aapo Kyrola, Andrew Tulloch, Yangqing Jia, and Kaiming He.
\newblock Accurate, large minibatch sgd: Training imagenet in 1 hour.
\newblock {\em arXiv preprint arXiv:1706.02677}, 2017.

\bibitem{gupta2019lvis}
Agrim Gupta, Piotr Dollar, and Ross Girshick.
\newblock {LVIS}: A dataset for large vocabulary instance segmentation.
\newblock In {\em CVPR}, 2019.

\bibitem{han2005borderline}
Hui Han, Wen-Yuan Wang, and Bing-Huan Mao.
\newblock Borderline-smote: a new over-sampling method in imbalanced data sets
  learning.
\newblock In {\em International conference on intelligent computing}, pages
  878--887. Springer, 2005.

\bibitem{he2017mask}
Kaiming He, Georgia Gkioxari, Piotr Doll{\'a}r, and Ross Girshick.
\newblock Mask r-cnn.
\newblock In {\em Proceedings of the IEEE international conference on computer
  vision}, pages 2961--2969, 2017.

\bibitem{he2016deep}
Kaiming He, Xiangyu Zhang, Shaoqing Ren, and Jian Sun.
\newblock Deep residual learning for image recognition.
\newblock In {\em Proceedings of the IEEE conference on computer vision and
  pattern recognition}, pages 770--778, 2016.

\bibitem{hu2018squeeze}
Jie Hu, Li Shen, and Gang Sun.
\newblock Squeeze-and-excitation networks.
\newblock In {\em Proceedings of the IEEE conference on computer vision and
  pattern recognition}, pages 7132--7141, 2018.

\bibitem{huang2016learning}
Chen Huang, Yining Li, Chen Change~Loy, and Xiaoou Tang.
\newblock Learning deep representation for imbalanced classification.
\newblock In {\em Proceedings of the IEEE conference on computer vision and
  pattern recognition}, pages 5375--5384, 2016.

\bibitem{huang2019mask}
Zhaojin Huang, Lichao Huang, Yongchao Gong, Chang Huang, and Xinggang Wang.
\newblock Mask scoring r-cnn.
\newblock In {\em Proceedings of the IEEE Conference on Computer Vision and
  Pattern Recognition}, pages 6409--6418, 2019.

\bibitem{kang2019decoupling}
Bingyi Kang, Saining Xie, Marcus Rohrbach, Zhicheng Yan, Albert Gordo, Jiashi
  Feng, and Yannis Kalantidis.
\newblock Decoupling representation and classifier for long-tailed recognition.
\newblock {\em arXiv preprint arXiv:1910.09217}, 2019.

\bibitem{kuznetsova2018open}
Alina Kuznetsova, Hassan Rom, Neil Alldrin, Jasper Uijlings, Ivan Krasin, Jordi
  Pont-Tuset, Shahab Kamali, Stefan Popov, Matteo Malloci, Tom Duerig, et~al.
\newblock The open images dataset v4: Unified image classification, object
  detection, and visual relationship detection at scale.
\newblock {\em arXiv preprint arXiv:1811.00982}, 2018.

\bibitem{lee2015deeply}
Chen-Yu Lee, Saining Xie, Patrick Gallagher, Zhengyou Zhang, and Zhuowen Tu.
\newblock Deeply-supervised nets.
\newblock In {\em Artificial intelligence and statistics}, pages 562--570,
  2015.

\bibitem{li2019gradient}
Buyu Li, Yu Liu, and Xiaogang Wang.
\newblock Gradient harmonized single-stage detector.
\newblock In {\em Proceedings of the AAAI Conference on Artificial
  Intelligence}, volume~33, pages 8577--8584, 2019.

\bibitem{lin2017feature}
Tsung-Yi Lin, Piotr Doll{\'a}r, Ross Girshick, Kaiming He, Bharath Hariharan,
  and Serge Belongie.
\newblock Feature pyramid networks for object detection.
\newblock In {\em Proceedings of the IEEE conference on computer vision and
  pattern recognition}, pages 2117--2125, 2017.

\bibitem{lin2017focal}
Tsung-Yi Lin, Priya Goyal, Ross Girshick, Kaiming He, and Piotr Doll{\'a}r.
\newblock Focal loss for dense object detection.
\newblock In {\em Proceedings of the IEEE international conference on computer
  vision}, pages 2980--2988, 2017.

\bibitem{lin2014coco}
Tsung-Yi Lin, Michael Maire, Serge Belongie, James Hays, Pietro Perona, Deva
  Ramanan, Piotr Doll{\'a}r, and C~Lawrence Zitnick.
\newblock Microsoft {COCO}: Common objects in context.
\newblock In {\em ECCV}, 2014.

\bibitem{liu2016ssd}
Wei Liu, Dragomir Anguelov, Dumitru Erhan, Christian Szegedy, Scott Reed,
  Cheng-Yang Fu, and Alexander~C Berg.
\newblock Ssd: Single shot multibox detector.
\newblock In {\em European conference on computer vision}, pages 21--37.
  Springer, 2016.

\bibitem{liu2019large}
Ziwei Liu, Zhongqi Miao, Xiaohang Zhan, Jiayun Wang, Boqing Gong, and Stella~X
  Yu.
\newblock Large-scale long-tailed recognition in an open world.
\newblock In {\em Proceedings of the IEEE Conference on Computer Vision and
  Pattern Recognition}, pages 2537--2546, 2019.

\bibitem{lu2019grid}
Xin Lu, Buyu Li, Yuxin Yue, Quanquan Li, and Junjie Yan.
\newblock Grid r-cnn.
\newblock In {\em Proceedings of the IEEE Conference on Computer Vision and
  Pattern Recognition}, pages 7363--7372, 2019.

\bibitem{mahajan2018exploring}
Dhruv Mahajan, Ross Girshick, Vignesh Ramanathan, Kaiming He, Manohar Paluri,
  Yixuan Li, Ashwin Bharambe, and Laurens van~der Maaten.
\newblock Exploring the limits of weakly supervised pretraining.
\newblock In {\em Proceedings of the European Conference on Computer Vision
  (ECCV)}, pages 181--196, 2018.

\bibitem{mikolov2013distributed}
Tomas Mikolov, Ilya Sutskever, Kai Chen, Greg~S Corrado, and Jeff Dean.
\newblock Distributed representations of words and phrases and their
  compositionality.
\newblock In {\em Advances in neural information processing systems}, pages
  3111--3119, 2013.

\bibitem{oh2016deep}
Hyun Oh~Song, Yu Xiang, Stefanie Jegelka, and Silvio Savarese.
\newblock Deep metric learning via lifted structured feature embedding.
\newblock In {\em Proceedings of the IEEE Conference on Computer Vision and
  Pattern Recognition}, pages 4004--4012, 2016.

\bibitem{peng2018megdet}
Chao Peng, Tete Xiao, Zeming Li, Yuning Jiang, Xiangyu Zhang, Kai Jia, Gang Yu,
  and Jian Sun.
\newblock Megdet: A large mini-batch object detector.
\newblock In {\em Proceedings of the IEEE Conference on Computer Vision and
  Pattern Recognition}, pages 6181--6189, 2018.

\bibitem{redmon2016you}
Joseph Redmon, Santosh Divvala, Ross Girshick, and Ali Farhadi.
\newblock You only look once: Unified, real-time object detection.
\newblock In {\em Proceedings of the IEEE conference on computer vision and
  pattern recognition}, pages 779--788, 2016.

\bibitem{ren2015faster}
Shaoqing Ren, Kaiming He, Ross Girshick, and Jian Sun.
\newblock Faster r-cnn: Towards real-time object detection with region proposal
  networks.
\newblock In {\em Advances in neural information processing systems}, pages
  91--99, 2015.

\bibitem{shen2016relay}
Li Shen, Zhouchen Lin, and Qingming Huang.
\newblock Relay backpropagation for effective learning of deep convolutional
  neural networks.
\newblock In {\em European conference on computer vision}, pages 467--482.
  Springer, 2016.

\bibitem{shrivastava2016training}
Abhinav Shrivastava, Abhinav Gupta, and Ross Girshick.
\newblock Training region-based object detectors with online hard example
  mining.
\newblock In {\em Proceedings of the IEEE conference on computer vision and
  pattern recognition}, pages 761--769, 2016.

\bibitem{shu2019meta}
Jun Shu, Qi Xie, Lixuan Yi, Qian Zhao, Sanping Zhou, Zongben Xu, and Deyu Meng.
\newblock Meta-weight-net: Learning an explicit mapping for sample weighting.
\newblock {\em arXiv preprint arXiv:1902.07379}, 2019.

\bibitem{wang2017learning}
Yu-Xiong Wang, Deva Ramanan, and Martial Hebert.
\newblock Learning to model the tail.
\newblock In {\em Advances in Neural Information Processing Systems}, pages
  7029--7039, 2017.

\bibitem{xie2017aggregated}
Saining Xie, Ross Girshick, Piotr Doll{\'a}r, Zhuowen Tu, and Kaiming He.
\newblock Aggregated residual transformations for deep neural networks.
\newblock In {\em Proceedings of the IEEE conference on computer vision and
  pattern recognition}, pages 1492--1500, 2017.

\bibitem{yin2019feature}
Xi Yin, Xiang Yu, Kihyuk Sohn, Xiaoming Liu, and Manmohan Chandraker.
\newblock Feature transfer learning for face recognition with under-represented
  data.
\newblock In {\em Proceedings of the IEEE Conference on Computer Vision and
  Pattern Recognition}, pages 5704--5713, 2019.

\bibitem{zhang2017range}
Xiao Zhang, Zhiyuan Fang, Yandong Wen, Zhifeng Li, and Yu Qiao.
\newblock Range loss for deep face recognition with long-tailed training data.
\newblock In {\em Proceedings of the IEEE International Conference on Computer
  Vision}, pages 5409--5418, 2017.

\end{thebibliography}
}

\end{document}